\documentclass[10pt,twocolumn,letterpaper]{article}

\usepackage[pagenumbers]{cvpr} 

\usepackage[dvipsnames]{xcolor}
\usepackage{tabularx, booktabs}
\usepackage{tikz}
\usepackage[accsupp]{axessibility}

\usepackage{tabularx}
\usepackage{booktabs}
\usepackage{adjustbox}
\usepackage{tabularx}
\usepackage{booktabs}
\usepackage{graphicx}
\usepackage{multirow}
\usepackage{indentfirst}
\usepackage{graphicx}
\usepackage{amsmath}
\usepackage{amssymb} 
\usepackage{array}
\usepackage{url}
\usepackage{multirow}
\usepackage{overpic}
\usepackage{amsfonts}

\usepackage{stfloats}

\definecolor{cvprblue}{rgb}{0.21,0.49,0.74}

\usepackage[pagebackref,breaklinks,colorlinks,citecolor=cvprblue]{hyperref}
\def\paperID{42} 
\def\confName{CVPR}
\def\confYear{2024}
\title{CRNet: A Detail-Preserving Network for Unified Image Restoration and Enhancement Task}
\author{Kangzhen Yang$^{1\dag}$ \ \ \ \ Tao Hu$^{1\dag}$ \ \ \ \ Kexin Dai$^{1}$ \ \ \ \ Genggeng Chen$^{2}$ \ \ \ \ Yu Cao$^{3}$ \ \ \\
\ \ Wei Dong$^{2}$ \ \ \ \ Peng Wu$^{1}$ \ \ \ \ Yanning Zhang$^{1}$\ \ \ \ Qingsen Yan$^{1}$\thanks{Corresponding author. $\dag$~The first two authors contributed equally to this work.
This work was partially supported by NSFC (62301432,62306240), NSBRPS (2023-JC-QN-0685, QCYRCXM-2023-057). 
}\\
$^{1}$Northwestern Polytechnical University\ 
$^{2}$Xi'an University of Architecture and Technology \\ $^{3}$Xi'an Institute of Optics and Precision Mechanics of CAS \\
\url{https://github.com/CalvinYang0/CRNet}}
\begin{document}
\maketitle
\begin{abstract}
\label{sec:Abstract}
In real-world scenarios, images captured often suffer from blurring, noise, and other forms of image degradation, and due to sensor limitations, people usually can only obtain low dynamic range images. To achieve high-quality images, researchers have attempted various image restoration and enhancement operations on photographs, including denoising, deblurring, and high dynamic range imaging. However, merely performing a single type of image enhancement still cannot yield satisfactory images. In this paper, to deal with the challenge above, we propose the Composite Refinement Network (CRNet) to address this issue using multiple exposure images. By fully integrating information-rich multiple exposure inputs, CRNet can perform unified image restoration and enhancement. To improve the quality of image details, CRNet explicitly separates and strengthens high and low-frequency information through pooling layers, using specially designed Multi-Branch Blocks for effective fusion of these frequencies. To increase the receptive field and fully integrate input features, CRNet employs the High-Frequency Enhancement Module, which includes large kernel convolutions and an inverted bottleneck ConvFFN. Our model secured third place in the first track of the Bracketing Image Restoration and Enhancement Challenge, surpassing previous SOTA models in both testing metrics and visual quality.
\end{abstract}    
\section{Introduction}
\label{sec:Introduction}

\begin{figure}
    \centering
    \includegraphics[width=1\linewidth]{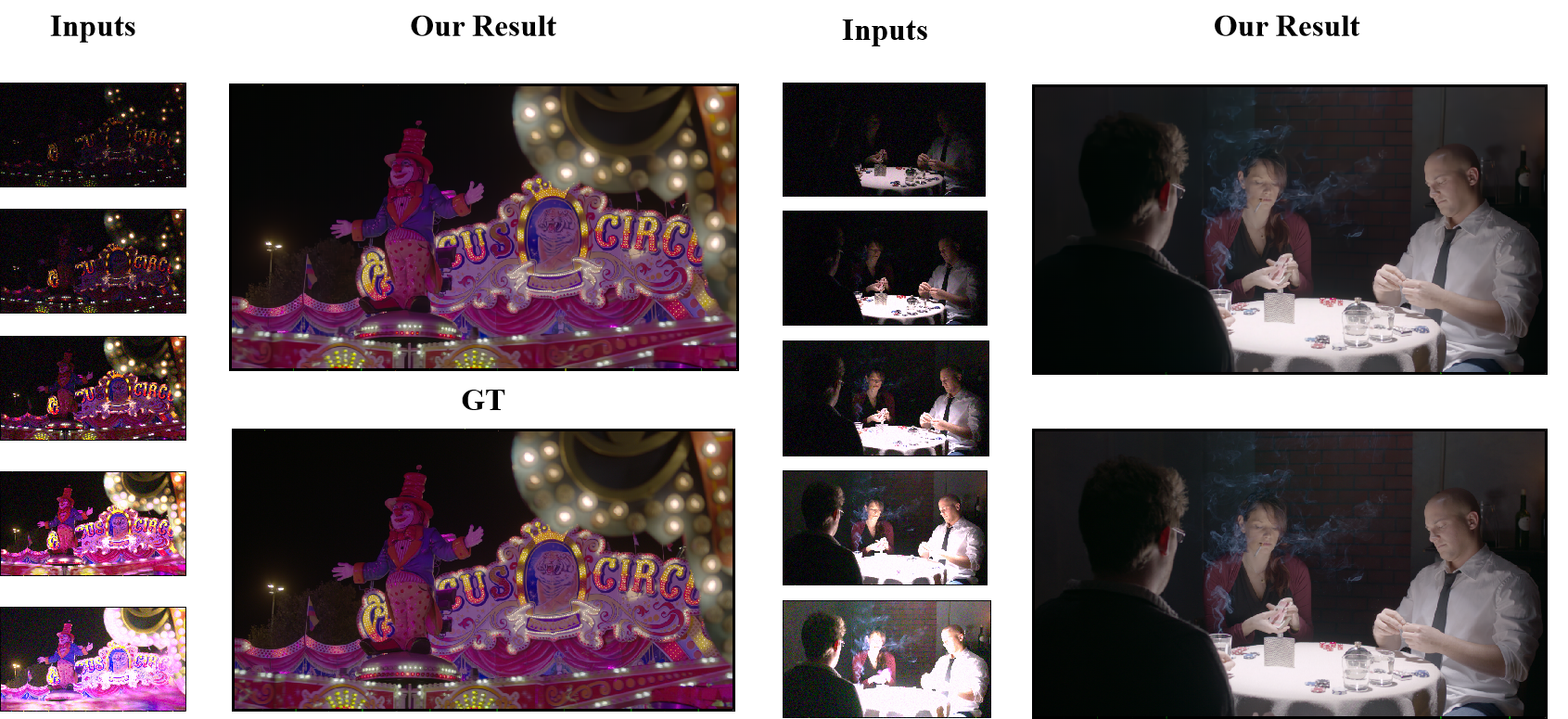}
    \caption{Our CRNet can finely restore the details lost in images from multiple exposure inputs and enhance them into HDR image.}
    \label{fig:Intro}
    \vspace{-0.7cm}
\end{figure}

In real photography scenarios, images captured are often subject to image degradation such as blur and noise due to lighting conditions and exposure time limitations. Moreover, due to typical sensor limitations, people usually obtain low dynamic range images. To address these issues, various methods have been studied for deblurring, denoising, and HDR imaging. However, these methods often focus only on individual specific tasks, and the resulting images are still unsatisfactory. To address this issue, establishing a model that can handle the Unified Image Restoration and Enhancement Task is crucial.

In previous studies, various single-image restoration methods have been proposed, such as denoising \cite{abdelhamed2020ntire,brooks2019unprocessing,guo2019toward,li2023ntire,zamir2020cycleisp,zhang2017beyond}, deblurring \cite{cho2021rethinking,mao2023intriguing,nah2017deep,tao2018scale,zamir2020cycleisp}, and high dynamic range reconstruction \cite{eilertsen2017hdr,lecouat2022high}. However, these methods often perform poorly due to limitations such as insufficient information from a single image and exposure time constraints. Consequently, people have started exploring methods for multi-exposure image restoration and enhancement. For instance, some studies \cite{chang2021low,lai2022face,shekarforoush2023dual,yuan2007image,zhang2022self} utilize dual-exposure images for image restoration, while others \cite{kalantari2017deep,liu2022ghost,niu2021hdr,prabhakar2020towards,yan2019attention,yan2020deep,yan2023towards,yan2022dual,yan2023smae,yan2023unified,hu2024generating} employ multi-exposure images to generate HDR images for image enhancement. These methods have made breakthroughs again, but they overlook the importance of edge and texture details in image restoration and enhancement tasks, lacking in the enhancement of high-frequency details. Recently, TMRNet\cite{zhang2024bracketing} provides a feasible solution for the unified image restoration and enhancement tasks. It takes multi-exposure images as input and progressively integrates non-reference frames with reference frames. However, with the deepening of the network and the increase in input images, such fusion methods lead to the gradual forgetting of earlier added frames by the network.

To address the aforementioned issues, we propose the Composite Refinement Network (CRNet) (Example visual effects are depicted in \cref{fig:Intro}). CRNet takes multi-exposure images as input and feeds them together into subsequent network parts after optical flow alignment, rather than frame by frame. To tackle the insufficient restoration of high-frequency details in image restoration tasks, CRNet explicitly employs pooling layers to separate high and low-frequency information for enhancement, and utilizes a specially designed non-stacked deep Multi-Branch Block for thorough fusion. To better integrate different image features, CRNet adopts the Convolutional Enhancement Block, a pure convolutional module primarily composed of large-kernel convolutions, which increases model's receptive field, along with ConvFFN featuring an inverted bottleneck structure to fully merge features. CRNet not only achieves breakthroughs in visual effects but also attains previous state-of-the-art performance in various evaluation metrics. Our CRNet secured third place in track 1 of the Bracketing Image Restoration and Enhancement Challenge. In summary, the main contributions are as follows:

\begin{itemize}
\item CRNet surpasses previous state-of-the-art (SOTA) models in both metrics and visual effects, achieving third place in track 1 of the Bracketing Image Restoration and Enhancement Challenge.

\item To address the issue of insufficient high-frequency details in image restoration tasks, we use pooling layers to explicitly separate high and low-frequency information and employ a Multi-Branch Block for fusion.
\item CRNet employs the Convolutional Enhancement Block, utilizing large kernel convolutions to increase the receptive field, along with ConvFFN featuring an inverted bottleneck structure for comprehensive feature fusion.
\end{itemize}

\section{Related Work}
\label{sec:Related work}
\paragraph{Burst Restoration and Enhancement}

Burst images refer to a series of images captured in rapid succession. During the capturing process, issues such as misalignment or blurring may arise due to camera movement or subject motion. The goal of burst image restoration and enhancement is to effectively use algorithms and techniques to process these consecutive images, in order to obtain high-quality, clear image outputs. Current methods\cite{wei2023towards,bhat2021deep,mehta2023gated,wu2020unpaired,godard2017deep,lucas1981iterative,mildenhall2018burst} typically involve techniques such as denoising, deblurring, image alignment, and super-resolution, with a primary focus on feature alignment and fusion.

In the aspect of denoising, Godard \etal \cite{godard2017deep} proposed a recursive fully convolutional deep neural network as a "feature accumulator" capable of processing images from a single frame to any number of burst frames. In terms of deblurring, Aittala \etal \cite{Aittala_2018_ECCV} introduced a novel convolutional architecture that can equally process information from all frames in a sequence, unaffected by their order. In the aspect of image enhancement, HDR+\cite{prabhakar2019fast} selects a frame with short exposure as the reference frame to avoid clipped highlights and motion blur, then aligns the other frames to this one before merging. It decides on a per-pixel basis whether to merge the image content or not. Building upon this, there has been work focused on addressing two or more tasks simultaneously. A method involving a deep reparametrization of the maximum a posteriori (MAP)\cite{bhat2021deep} formulation was proposed, which involves reparametrizing the classical MAP into a deep feature space. This approach shows good performance on both super-resolution and burst denoising tasks.

\paragraph{Multi-Exposure HDR Image Reconstruction.}

HDR imaging refers to the process of reconstructing a High Dynamic Range (HDR) image from a series of Low Dynamic Range (LDR) images with different exposures. A key challenge in this task is how to align multiple input frames to address the ghosting problem. Traditional alignment methods primarily utilize alignment-based\cite{bogoni2000extending,kang2003high,tomaszewska2007image}, rejection-based\cite{grosch2006fast,pece2010bitmap,zhang2011gradient}, and patch-based methods\cite{sen2012robust,hu2013hdr,ma2017robust} for alignment. Although they have achieved certain breakthroughs, they often err under conditions of extreme lighting and motion. The emergence of deep learning has provided new ideas for this issue. Some researchers have used CNN-based methods for alignment\cite{kalantari2017deep,yan2019attention,yan2020deep,niu2021hdr,9826814}. Kalantari \etal \cite{kalantari2017deep} first aligned images using optical flow and then fused them through a CNN network. Yan \etal \cite{yan2019attention} introduced a CNN-based spatial attention mechanism to suppress movement and oversaturated areas. Yan \etal \cite{yan2020deep} designed a non-local module to expand the receptive field for global merging. CNN-based alignment methods have made further progress, but still result in ghosting in overexposed areas due to motion. The advent of Transformers has given researchers new ideas for alignment methods. Song \etal \cite{song2022selective} utilized the large receptive field of the Transformer to globally recover motion areas. 
\section{Proposed Method}
\label{sec:Proposed method}

\begin{figure*}
    \centering
    \includegraphics[width=1\linewidth,height=0.35\textwidth]{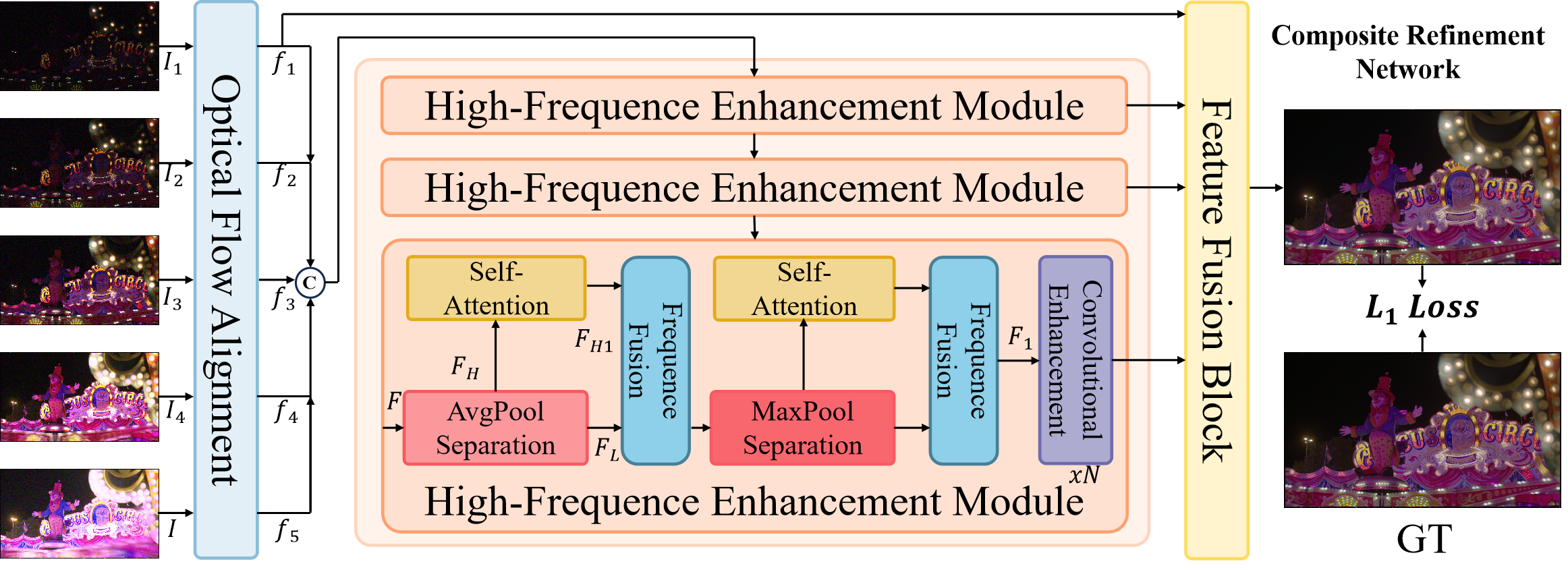}
    \caption{An overview of the CRNet model reveals that it mainly comprises three components: alignment, high-frequency enhancement, and output fusion. In the High-Frequency Enhancement Module, we swiftly separate high and low-frequency features and then utilize  cleverly designed Multi-Branch Block to fuse them. Subsequently, we employ a purely Convolutional Enhancement Block to efficiently extract and fuse features for image enhancement.}
    \label{fig:Overview of CRNet}
    \vspace{-0.6cm}
\end{figure*}

We are given a sequence of Raw images $\{R_1, R_2, \ldots, R_N\}$ captured in a dynamic scene, each with a distinct exposure level, and our objective includes simultaneous denoising, deblurring, and HDR reconstruction. The resultant image should closely mirror a reference image, denoted as $R_r$. Specifically, five Raw images $\{R_1, R_2, R_3, R_4, R_5\}$, ranked by their exposure from lowest to highest, are selected for input, with $R_1$ serving as the reference. Then, we normalize $R_i$ to $\frac{R_i}{\Delta t_i / \Delta t_1}$, and the term \(\Delta t_i\) refers to the exposure time of the \( i \)-th image.
Following the recommendations from multi-exposure HDR reconstruction methods \cite{liu2022ghost,niu2021hdr,tel2023alignment,yan2019attention,yan2023unified}, we transform the raw images after normalizing with gamma mapping to obtain $\{L_1, L_2, L_3, L_4, L_5\}$:
\begin{equation}
L_{i} = \left( \frac{R_{i}}{\Delta t_{i} / \Delta t_{1}} \right)^{\gamma},
\end{equation}
where \(\gamma\) represents the gamma correction parameter and is generally set to $\frac{1}{2.2}$, the same as TMRNet\cite{zhang2024bracketing} does in its code. 

Subsequently, we concatenate each $L_i$ with its corresponding $R_i$ to form $\{I_1, I_2, I_3, I_4, I_5\}$, according to the following equation:
\begin{equation}
I_{i} = \left\{ \frac{R_{i}}{\Delta t_{i} / \Delta t_{1}}, L_{i} \right\},
\end{equation}

Then, we concatenate each $L_i$ with its corresponding $R_i$ to form $\{I_1, I_2, I_3, I_4, I_5\}$. Feeding $\{I_1, I_2, I_3, I_4, I_5\}$ into our model, we derive an HDR image free of noise and blur, denoted as $\hat{H}$, according to the following equation:
\begin{equation}
\hat{H} = f (I_1, I_2, I_3, I_4, I_5; \theta),
\end{equation}
where the function $f (\cdot)$ signifies the imaging network, and $\theta$ encapsulates the parameters of the network.

\subsection{Overview of the CRNet}

As shown in \cref{fig:Overview of CRNet}, within CRNet, for the inputs $\{I_1, I_2, I_3, I_4, I_5\}$, we first align five input images to get $\{f_1, f_2, f_3, f_4, f_5\}$, employing an Optical Flow Alignment Block to align the five input images, which includes convolutional blocks for shallow feature extraction and a pre-trained Spynet\cite{ranjan2017optical}. Experiments(See in See the experimental section, "The Structure to Fuse Features" part) indicate that the frame-by-frame input method adopted by TMRNet\cite{zhang2024bracketing} could lead to the network gradually forgetting earlier added images as the network deepens and the number of input images increases, shifting focus to more recently added images and thus degrading image quality. Therefore, we concatenate the five aligned images as the input for subsequent processing, allowing for more comprehensive utilization of the information from each image. 

Subsequently, the merged features are sequentially passed through 3 High-Frequency Enhancement Modules. Each module initially utilizes two distinct pooling layers to separate and individually amplify high and low-frequency information\cite{lu2022transformer}, followed by \(N\) Convolutional Enhancement Blocks. After frequency separation, the valuable high-frequency information is enhanced using Transformer, and the information across frequencies is fused using Multi-Branch Blocks\cite{tian2024multi}, thereby fully restoring the details of the image. Convolutional Enhancement Blocks can be regarded as high-frequency filters, employing large-kernel depth-wise separable convolutions \cite{liu2022convnet} and convolutionalized FFN \cite{ding2022scaling} to increase the receptive field and fully merge features, while further enhancing high-frequency information.

Finally, the outcomes of the 3 High-Frequency Enhancement Sub-networks are fused with the reference frame, which has been aligned using simple convolutional blocks, to output the final result.

\subsubsection{Frequence Separation and Fusion}
\begin{figure}
    \centering
    \includegraphics[width=1\linewidth]{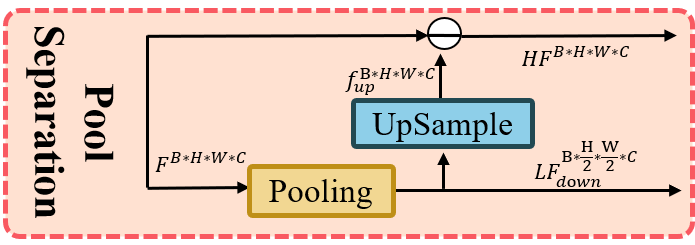}
    \caption{We efficiently separate high and low-frequency information through simple pooling layers.}
    \label{fig:Frequence Fusion}
    \vspace{-0.4cm}
\end{figure}

\begin{figure}
    \centering
    \includegraphics[width=1\linewidth]{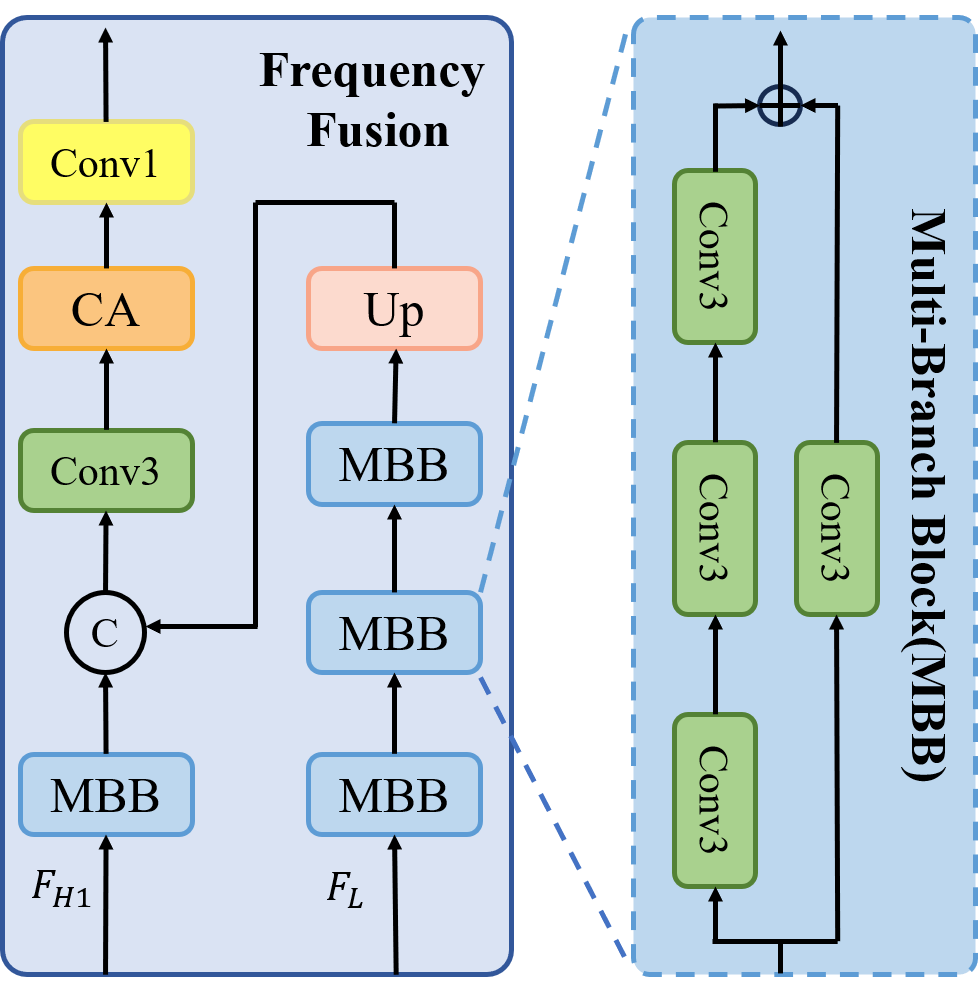}
    \caption{Through asymmetric parallel convolutional groups, the model effectively integrates high and low-frequency information.}
    \label{fig:Multi-Branch Block}
    \vspace{-0.5cm}
\end{figure}

In the unified image restoration and enhancement task, enhancing high-frequency feature information is crucial, and one effective method is to separately amplify high-frequency and low-frequency features. Traditional separation techniques, such as Fourier transforms, entail significant computational overhead and are not straightforwardly integrated into networks. To minimize the computational expenses associated with segregating high and low-frequency features, we employ pooling layers\cite{lu2022transformer} for this task (as shown in the \cref{fig:Multi-Branch Block}). In the CRNet, we specifically utilized average pooling and max pooling. Specifically, a pooling layer down-samples the input feature maps with dimensions \( B \times H \times W \times C \) to obtain the low-frequency features \( F_L \) at a reduced resolution \( F_L^{B \times \frac{H}{2} \times \frac{W}{2} \times C} \). Bilinear interpolation is then used to upsample these features back to the original dimensions, \( f_{\text{up}}^{B \times H \times W \times C} \). The high-frequency features \( F_H \) of \( F \) are subsequently computed by subtracting \( F_L\) from the original features \( F \). This approach enables us to quickly capture the image's high and low-frequency characteristics, as described by the equation below:
\begin{equation}
F_L = \text{Pooling}(F),
\end{equation}
\begin{equation}
F_H = F - \text{Upsample}(F_L),
\end{equation}
where \text{Pooling} refers to the down-sampling operation using a pooling layer, and \text{Upsample} refers to the up-sampling operation using bilinear interpolation.

After obtaining the explicitly extracted high-frequency features \( F_H \), we opt to globally enhance them using Self-Attention mechanism to fully exploit the high-frequency information and get \( F_{H_1}\). However, simply fusing features across different frequency domains through basic convolutional blocks could lead to information loss and insufficient fusion. Therefore, to adequately merge high-frequency and low-frequency information, we primarily utilize the Multi-Branch Block (See in \cref{fig:Multi-Branch Block}). 

The Multi-Branch Block consists of a dual-path convolutional assembly, with each path employing a different number of convolutions. The first branch, equipped with three convolutional kernels, focuses more on the high-frequency details of the image, while the second branch, with a single convolutional kernel, pays more attention to the low-frequency content and contours of the image\cite{tian2024multi}. This allows our network to effectively meet the requirements for merging high and low-frequency features. 

Overall, we first pass the high-frequency and low-frequency features through several Multi-Branch Blocks individually, then upsample the low-frequency information to match the dimensions of the high-frequency features, concatenate them together, and finally fuse them through channel attention and simple convolution. The specific formulas are as follows:

\begin{equation}
H = \text{MBB}(F_{H1}),
\end{equation}
\begin{equation}
L = \text{MBB}(\text{MBB}(\text{MBB}(F_L))),
\end{equation}
\begin{equation}
\text{Out} = \text{Conv}_{1 \times 1}(\text{CA}(\text{Conv}_{3 \times 3}(\text{Concatenate}(\text{Up}(L), H)))),
\end{equation}
where $\text{MBB}$ represents the Multi-Branch Block, $\text{Conv}_{1 \times 1}$ represents a $1 \times 1$ convolution, $\text{Conv}_{3 \times 3}$ represents a $3 \times 3$ convolution, $\text{CA}$ represents channel attention, and $\text{Up}$ represents bilinear interpolation upsampling.

\subsubsection{Convolutional Enhancement}
\begin{figure}
    \centering
    \includegraphics[width=1\linewidth,height=0.2\textwidth]{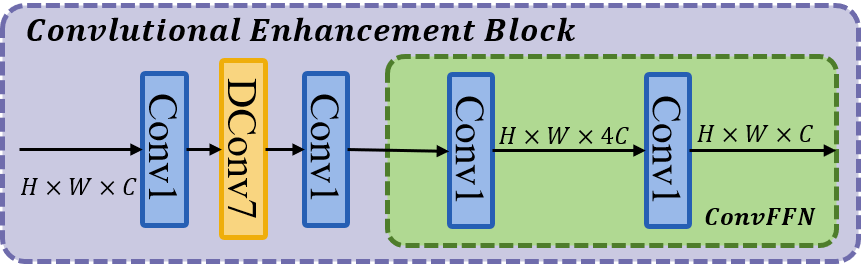}
    \caption{Our Convolutional Enhancement Block utilizes depth-wise separable convolutions with large kernels to achieve a large receptive field. Additionally, it leverages ConvFFN with an inverted bottleneck structure to enhance feature fusion capability.}
    \label{fig:Convolutional Enhancement Block}
    \vspace{-0.5cm}
\end{figure}
To improve the receptive field and thoroughly integrate the features from inputs, we employ the Convolutional Enhancement Block in our network (see \cref{fig:Convolutional Enhancement Block}). This purely convolutional module utilizes 7$\times$7 depth-wise separable convolutions to achieve a broad receptive field and features an inverted bottleneck structure ConvFFN for thorough information extraction. Additionally, it acts as a high-pass filter, facilitating efficient content fusion from the five input frames and implicitly enhancing high-frequency information. The formula for the entire process is as follows:
\begin{equation}
F_2 = \text{ConvFFN}(\text{Conv}_{1\times 1}(\text{DConv}_{7\times 7}(\text{Conv}_{1\times 1}(F_1))))
\end{equation}
Here, \(\text{Conv1}_{1\times 1}\) represents a 1$\times$1 convolution, \(\text{DConv}_{7\times 7}\) denotes depth-wise separable 7$\times$7 convolution, and each convolution layer is followed by GELU activation for non-linear activation.

\section{Experiments}
\label{sec:Experiments}

\subsection{Experiments Settings}
\textbf{Datasets.}
The dataset we utilized is the training set from track 1 of the Bracketing Image Restoration challenge. This dataset is constructed based on HDR videos to simulate the motion and blur effects of multi-exposure images. Initially, HDR videos from Froehlich \etal \cite{froehlich2014creating} were extracted and processed by frame interpolation and conversion into Bayer-pattern HDR raw sequences. Subsequently, degradation operations were introduced to generate multi-exposure images, including downsampling, grouping, blur synthesis, conversion to LDR images, and noise addition. Finally, the dataset comprises 1355 data pairs, with 1045 pairs used for training and 290 pairs for testing.

\textbf{Training Loss.}
Following the approach of TMRNet, we employ the $\mu$-law to map the resulting image from the linear domain to the tone-mapping domain:
\begin{equation}
T(x) = \frac{\log(1 + \mu x)}{\log(1 + \mu)},
\end{equation}
where $\mu$=5000.

Given the result $\hat{H}$ from CRNet and the ground truth $H$, we conduct pixel-level L1 loss in the tone-mapping domain as follows:
\begin{equation}
L = \lVert T(H) - T(\hat{H}) \rVert_1,
\end{equation}

\textbf{Evaluation Metrics.}
We compute five commonly used metrics for evaluation, namely PSNR-$\mu$ and SSIM-$\mu$, where 'L' denotes the linear domain, and '$\mu$' denotes the tone-mapping domain. In the $\mu$-law tone-mapping, $\mu = 5000$. In calculating the PSNR metric, the maximum value is set to 1. In the calculation of the SSIM metric, the maximum dynamic range is 1.

\textbf{Implementation Details.}
During the training process, images are randomly cropped to patches with a size of \(128 \times 128\), and are subject to random flipping and rotation. In CRNet, \(N\) is set to 10. We used the PyTorch framework and adopted the AdamW optimizer with $\beta_1$ = 0.9 and $\beta_2$ = 0.999. The initial learning rate was set to $10^{-4}$, using StepLR for learning rate decay by multiplying it by 0.5 every 80 epochs. The model was trained on the synthetic dataset provided by the BracketIRE Task on 4 A100 GPUs for a total of 500 epochs, amounting to 3 days of training. We utilized Hugging Face's Accelerator for parallel training.

\subsection{Comparison with the State-of-the-art Methods}
\begin{table}[h]
    \centering
    \footnotesize
    \setlength{\tabcolsep}{1pt} 
    \caption{The evaluation results on the Bracketing Image Restoration and Enhancement Challenge\cite{zhang2024bracketing}'s dataset for Track1 and Track2. The best and second-best results are highlighted in \textbf{Bold} and \underline{Underline}, respectively.}
    \label{tab:quantitative results}
    \textbf{Track1 Results}
    \begin{tabular}{c|cccccccc}
        \toprule
        \textbf{Models} & \textbf{AHDRNet} & \textbf{CA-VIT} & \textbf{HyHDR} & \textbf{SCTNet} & \textbf{TMRNet} & \textbf{Kim's} & \textbf{Ours} \\
        \midrule
        PSNR-$\mu$ & 36.32 & 36.54 & 37.4 & 36.90 & \underline{38.19} & 37.93 & \textbf{39.03} \\
        SSIM-$\mu$ & 0.927 & 0.9341 & 0.940 & 0.944 & \underline{0.949} & 0.945 & \textbf{0.950} \\
        \bottomrule
    \end{tabular}
    
    \vspace{3pt}
    \textbf{Track2 Results}
    \vspace{3pt} 
    \begin{tabular}{c|cccccccc}
        \toprule
        \textbf{Models} & \textbf{AHDRNet} & \textbf{CA-VIT} & \textbf{HyHDR} & \textbf{SCTNet} & \textbf{TMRNet} & \textbf{Kim's} & \textbf{Ours} \\
        \midrule
        PSNR-$\mu$ & 28.17 & 28.18 & 28.59 & 28.28 & \underline{28.91} & 28.33 & \textbf{29.45} \\
        SSIM-$\mu$ & 0.842 & 0.848 & 0.849 & 0.846 & \underline{0.857} & 0.849 & \textbf{0.871} \\
        \bottomrule
    \end{tabular}
    \vspace{-0.3cm}
\end{table}

\begin{figure*}[h]
  \centering

  \subfloat[]{%
    \includegraphics[width=1\linewidth]{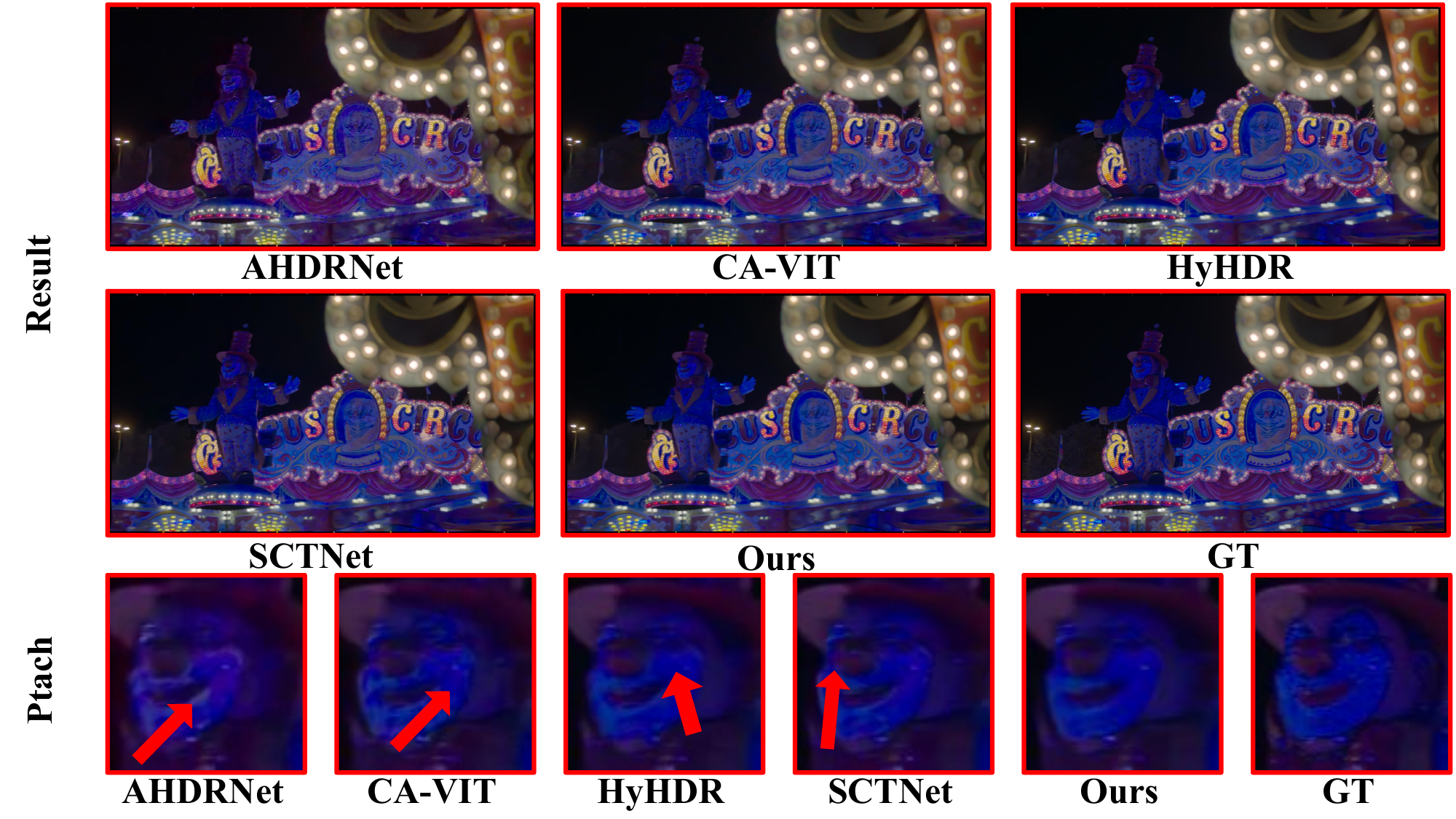}%

  }

  \hfill

  \subfloat[]{%
    \includegraphics[width=1\linewidth]{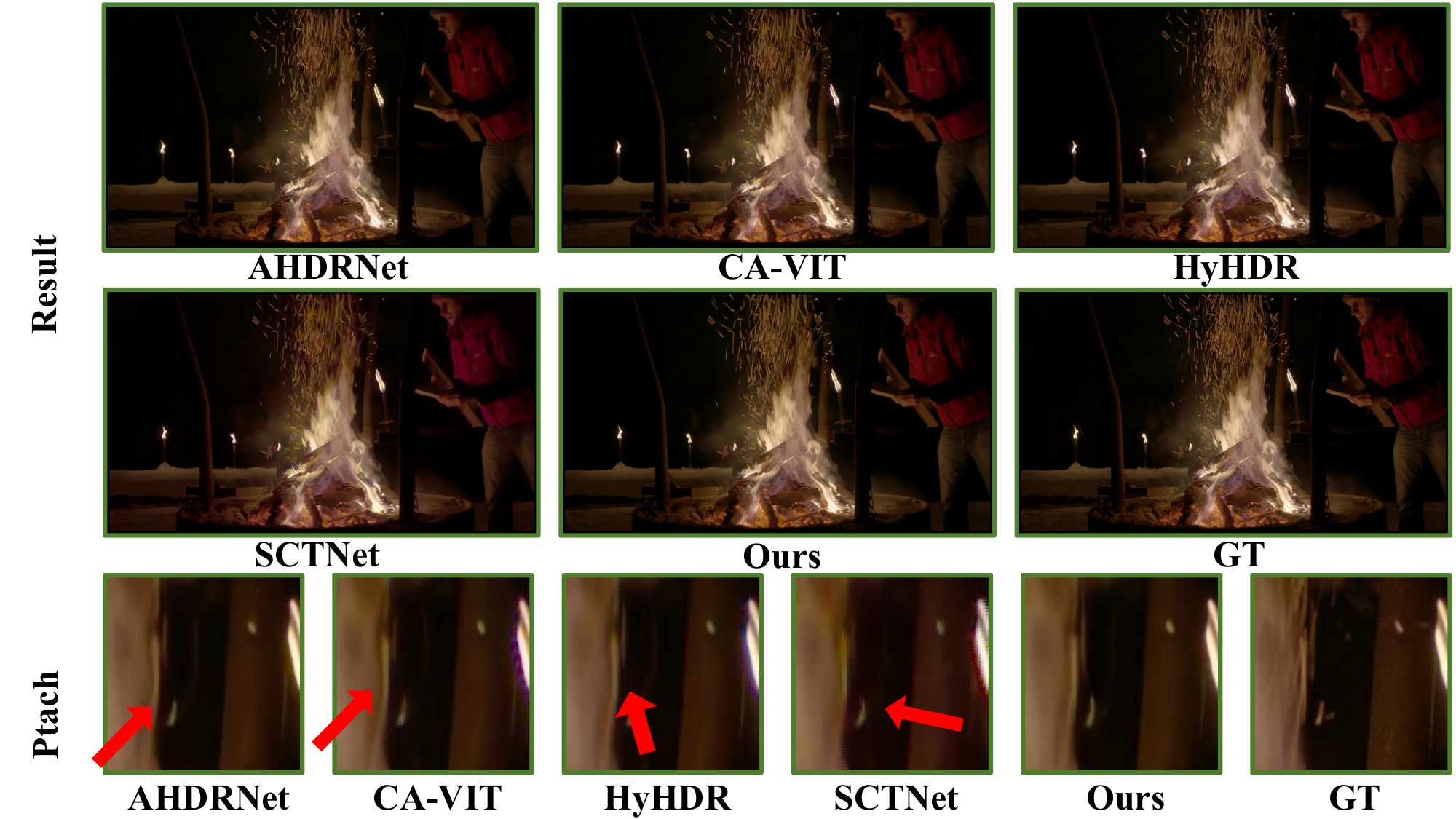}%

  }

  \caption{Examples of comparisions on the track 1 of the Bracketing Image Restoration and Enhancement Challenge dataset.}
  \label{fig:comparison}
\end{figure*}
To further validate the superiority of our model, we compared CRNet with several state-of-the-art deep learning models AHDRNet \cite{yan2019attention}, TMRNet\cite{zhang2024bracketing} Transformer-based CA-ViT \cite{liu2022ghost}, SCTNet\cite{tel2023alignment}, Kim's\cite{kim2023joint} and HyHDR \cite{yan2023unified} quantitatively and qualitatively on track 1 of the Bracketing Image Restoration and Enhancement Challenge dataset. As shown in \cref{tab:quantitative results}, our approach outperforms the compared models in the testing metrics. From \cref{fig:comparison}(a), it can be observed that in the bonfire scene, where the background color is too dark and the flame area is too bright, and the fluttering bonfire presents foreground object motion, which is extremely challenging. Thanks to the proposed high-frequency enhancement module, our model successfully delineates clear and distinct edges in the flame's boundary region and suppresses ghosting, while other models exhibit edge discontinuities or redundancies. Additionally, in the facial area (see in \cref{fig:comparison}(b)), the facial contour, especially the nose area, is particularly challenging. Our CRNet successfully restores the contour of the nose, whereas other models lack contour on the left side of the nose or exhibit unclear delineation on the right side. This indicates a significant improvement of our CRNet over previous methods, especially in scenes with too dark or too bright conditions, in high-frequency detail regions, and also surpasses previous state-of-the-art methods in numerical metrics. 
Furthermore, we introduced an upsampling module similar to that of TMRNet into the base of CRNet and conducted tests on track 2 of the Bracketing Image Restoration and Enhancement Challenge dataset. As shown in \cref{tab:quantitative results}, our approach outperforms the compared models in the testing metrics.

\subsection{Results of track 1 of the Bracketing Image Restoration and Enhancement Challenge}
\begin{table*}[t!]
    \centering
    \normalsize
    \caption{Results\cite{zhang2024ntirebrack} on track 1 of the Bracketing Image Restoration and Enhancement Challenge. Quantitative metrics are calculated on the full images. \#FLOPs, inference time, and GPU memory are measured when generating a $1920\times1080$ raw image. The test uses NVIDIA RTX A6000 GPU to calculate the inference time and adopts the THOP~\cite{thop} tool to calculate \#FLOPs. The ranking is based on the PSNR metric of the full images. The best and second-best results are highlighted in \textbf{Bold} and \underline{Underline}, respectively.}

    \label{tab:track1results}
    \begin{tabular}{ccccccc}
    \toprule
    Rank & Team & Full Images PSNR $\uparrow$ & \#Params (M) & \#FLOPs (T) & Time (s) & Memory (GB) \\
    \midrule
    1 & SRC-B & 40.54 & 94.34 & 48.238 & 3.102 & 20 \\
    2 & MegIRE & 39.78 & 19.75 & 30.751 & 2.383 & 16 \\
    3 & UPN1(CRNet) & 39.03 & 13.32 & \textbf{10.409} & \textbf{1.090} & \textbf{6} \\
    4 & CVG & 38.78 & 13.29 & \underline{21.340} & 7.518 & \underline{11} \\
    5 & FZU\_DXW & 38.46 & 14.04 & 22.283 & \underline{1.829} & 16 \\
    \midrule
    - & TMRNet~\cite{zhang2024bracketing} & 38.19 & 13.29 & 21.340 & 1.874 & 15 \\
    \bottomrule
    \end{tabular}
\vspace{-0.3cm}  
\end{table*}

As shown in \cref{tab:track1results}, we achieved the third place in track 1 of the Bracketing Image Restoration and Enhancement Challenge. Compared to other ranking models, our CRNet significantly outperforms in terms of FLOPS, inference time, and GPU memory usage.computational costs.
\subsection{Ablation Studies}
\begin{table}[h]
\caption{The Ablation study of CRNet.'-' indicates that a certain structure is not applicable, and another module with similar computational complexity is embedded to replace it.}
\centering
\begin{tabular}{l|cc}
\hline
\textbf{Models} & \textbf{PSNR-$\mu$} & \textbf{SSIM-$\mu$} \\
\hline

CRNet & \textbf{39.03} & \textbf{0.949} \\
\hline
CRNet-Frequency Separation & 38.71 & 0.922 \\
\hline
CRNet-Multi-Branch Block & 38.64 & 0.937 \\
\hline
CRNet-Large Kernel Conv & 38.76 & 0.941 \\
\hline
CRNet-ConvFFN & 38.69 & 0.933 \\
\hline
CRNet+Recurrent Structure & 38.79 & 0.945\\
\hline
\end{tabular}
\label{tab:Ablation for CRNet}
\end{table}

To validate the effectiveness of each module in our CRNet, we conducted a series of ablation experiments(See in \cref{tab:Ablation for CRNet}):
\begin{itemize}
\item \textbf{Model1:} We use the full CRNet.
\item \textbf{Model2:} We do not perform frequency separation; instead, we directly pass the original feature maps through self-attention and merge them with the results after downsampling.

\item \textbf{Model3:} In the frequence fusion part, we do not utilize the Multi-Branch Block structure; instead, we simplify it to four consecutive convolutional layers.
\item \textbf{Model4:}We replace the $7 \times 7$ convolution in the Convolutional Enhancement with three $3 \times 3$ convolutions.
\item \textbf{Model5:} We do not employ the inverted bottleneck structure in ConvFFN. Instead, we use two $1 \times 1$ convolutions without changing the number of channels.
\item \textbf{Model6:} While keeping the model's parameter count almost identical, we arrange all the High-Frequency Enhancement Modules in CRNet according to the recurrent structure of TMRNet, sequentially feeding images into them. The output of the last High-Frequency Enhancement Module is then combined with the reference frame and input into the Feature Fusion Block.
\end{itemize}

\textbf{Frequence Separation.}

\begin{figure}
    \centering
    \includegraphics[width=1\linewidth]{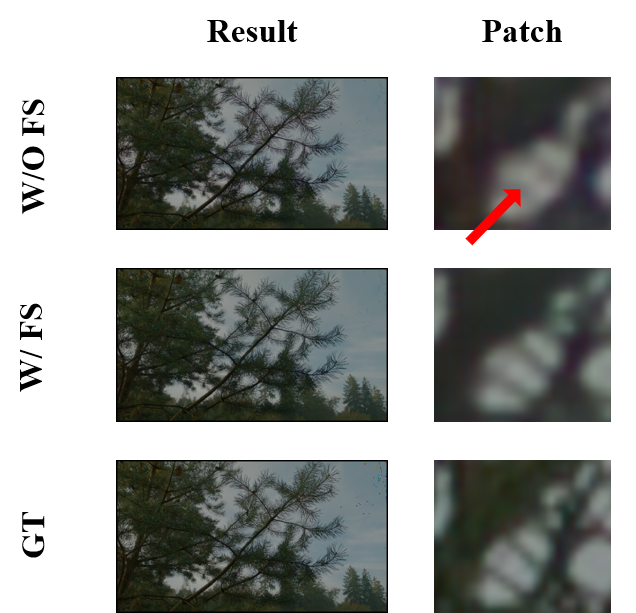}
    \caption{From the figure, it can be observed that our CRNet successfully restores the edges of the branches clearly after using frequency separation.\textbf{FS} represents for Frequence Separation Block.}

    \label{fig:FS Extra}
    \vspace{-0.2cm}
\end{figure}

As shown in Table~\ref{tab:Ablation for CRNet}, Model 1 outperforms Model 2 by 0.32 in PSNR-$\mu$ and 0.027 in SSIM-$\mu$. This indicates that utilizing frequency separation to extract high-frequency features and enhancing them selectively significantly improves the image quality. As illustrated in \cref{fig:FS Extra}, after employing frequency separation, our CRNet successfully restores the edges of the branches clearly after using frequency separation.

\textbf{Multi-Branch Block.}
\begin{figure}
    \centering
    \includegraphics[width=1\linewidth]{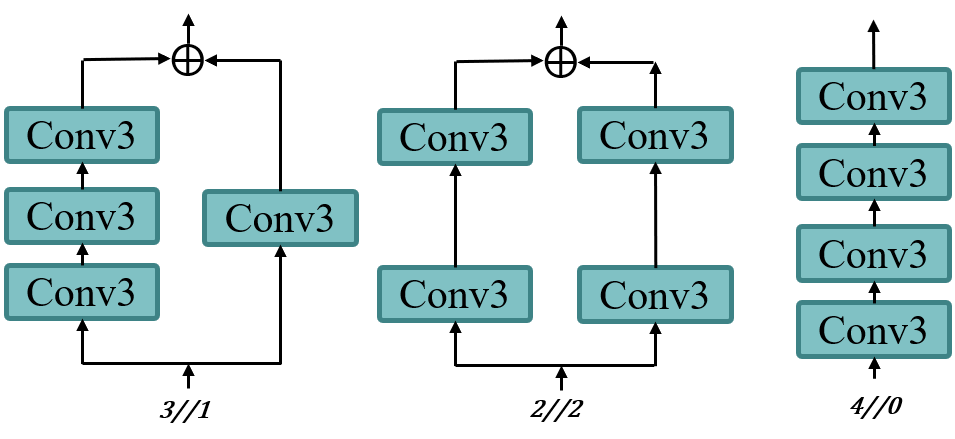}
    \caption{"\textbackslash\textbackslash" represents parallel inputs. As shown in the figure, we attempted three different ways of allocating convolutional blocks.}
    \label{fig:Ablation for Multi-Branch Block Structure}
\vspace{-0.5cm}
\end{figure}
\begin{table}[h]
\caption{Ablation for Multi-Branch Block. The allocation scheme of having three convolution blocks on one side and one on the other side appears to be the most effective.}
\centering 
\begin{tabular}{l|cc}
\hline
\textbf{Models} & \textbf{PSNR-$\mu$} & \textbf{SSIM-$\mu$} \\
\hline
3\textbackslash\textbackslash1 & \textbf{39.03} & \textbf{0.949} \\
\hline
2\textbackslash\textbackslash2 & 38.82 & 0.941 \\
\hline
4\textbackslash\textbackslash0 & 38.64 & 0.937 \\
\hline
\end{tabular}
\vspace{-0.3cm}
\label{tab:Ablation for Multi-Branch Block}
\end{table}
As shown in Table~\ref{tab:Ablation for CRNet}, Model 1 outperforms Model 3 by 0.39 in PSNR-$\mu$ and 0.012 in SSIM-$\mu$. This indicates that our utilization of the Multi-Branch Block for the fusion of high and low-frequency content is highly effective. To further validate the effectiveness of our structure, we conducted more detailed ablation experiments. As illustrated in \cref{fig:Ablation for Multi-Branch Block Structure}, we reconfigured the positions of the four convolutional kernels for investigation. As shown in \cref{tab:Ablation for Multi-Branch Block}, Model 1 performs the best, implying that our Multi-Branch Block structure achieves better performance under the same computational cost.

\textbf{Convolutional Enhancement Block.}

As shown in Table~\ref{tab:Ablation for CRNet}, Model 1 outperforms Model 4 by 0.27 in PSNR-$\mu$ and 0.008 in SSIM-$\mu$. Model 1 outperforms Model 5 by 0.34 in PSNR-$\mu$ and 0.016 in SSIM-$\mu$. This indicates that our Convolutional Enhancement Block's large kernel convolution design and ConvFFN fusion of input features are highly effective. 

\begin{figure}
    \centering
    \includegraphics[width=1\linewidth]{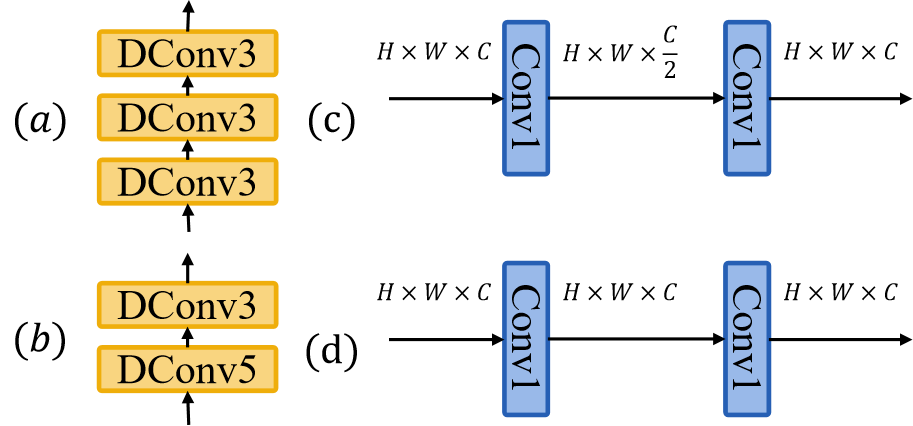}
    \caption{The figure shows the model diagrams of the comparison models included in \cref{tab:Ablation for CEB}}
    \label{fig:extra ceb}
    \vspace{-0.5cm}
\end{figure}
\begin{table}[h]
\caption{The stuctures presented in the table are depicted in \cref{fig:extra ceb}. Models a and b are used to replace the \(7 \times 7\) depthwise separable convolution, while models c and d are used to replace the inverse bottleneck ConvFFN.
}
\centering 
\begin{tabular}{l|cc}
\hline
\textbf{Models} & \textbf{PSNR-$\mu$} & \textbf{SSIM-$\mu$} \\
\hline
CRNet & \textbf{39.03} & \textbf{0.949} \\
\hline
$3 \times 3 \times 3$(a) & 38.76 & 0.941 \\
\hline
$5 \times 5$+$3 \times 3$(b) & 38.88 & 0.939 \\
\hline
normal bottleneck(c) & 38.69 & 0.933 \\
\hline
normal \(1 \times 1\) convolutions(d) & 38.82 & 0.945 \\
\hline
\end{tabular}
\label{tab:Ablation for CEB}
\vspace{-0.4cm}
\end{table}
To validate the effectiveness of our Convolutional Enhancement Block structure, we conducted further ablation experiments. (See in \cref{fig:extra ceb})We attempted to replace the \(7 \times 7\) depthwise separable convolution with either three \(3 \times 3\) depthwise separable convolutions or a stack of one \(5 \times 5\) convolution followed by one \(3 \times 3\) convolution, and we tried to replace the ConvFFN with a normal bottleneck structure or a normal \(1 \times 1\) convolution chaining structure. As shown in \cref{tab:Ablation for CEB}, the experimental results indicate that the \(7 \times 7\) depthwise separable convolution and the inverse bottleneck ConvFFN adopted by us are more optimal structures for the tasks of image restoration and enhancement.

\textbf{The Structure to fuse features}
As shown in Table~\ref{tab:Ablation for CRNet}, Model 1 outperforms Model 6 by 0.24 in PSNR-$\mu$ and 0.005 in SSIM-$\mu$. From this, it is evident that inputting the feature maps together into the High-Frequency Enhancement Module is a better approach than using the Recurrent Structure.
\section{Conclusion}
\label{sec:Conclusion}
We introduced the Composite Refinement Network (CRNet), which uniformly completes image restoration and enhancement. CRNet explicitly separates the frequency domain and employs the Multi-Branch Block for targeted fusion, thereby enhancing high-frequency details in image restoration. Additionally, it utilizes the Convolutional Enhancement Block, leveraging large kernel convolutions and ConvFFN to increase the receptive field and enhance feature fusion capabilities. CRNet surpassed previous state-of-the-art models in both metrics and visual effects, and achieved third place in track 1 of the Bracketing Image Restoration and Enhancement Challenge, with computational costs far below those of other models.

{
    \small
    \bibliographystyle{ieeenat_fullname}
    \bibliography{main}

\begin{thebibliography}{59}
\providecommand{\natexlab}[1]{#1}
\providecommand{\url}[1]{\texttt{#1}}
\expandafter\ifx\csname urlstyle\endcsname\relax
  \providecommand{\doi}[1]{doi: #1}\else
  \providecommand{\doi}{doi: \begingroup \urlstyle{rm}\Url}\fi

\bibitem[Abdelhamed et~al.(2020)Abdelhamed, Afifi, Timofte, and Brown]{abdelhamed2020ntire}
Abdelrahman Abdelhamed, Mahmoud Afifi, Radu Timofte, and Michael~S Brown.
\newblock Ntire 2020 challenge on real image denoising: Dataset, methods and results.
\newblock In \emph{Proceedings of the IEEE/CVF Conference on Computer Vision and Pattern Recognition Workshops}, pages 496--497, 2020.

\bibitem[Aittala and Durand(2018)]{Aittala_2018_ECCV}
Miika Aittala and Fredo Durand.
\newblock Burst image deblurring using permutation invariant convolutional neural networks.
\newblock In \emph{The European Conference on Computer Vision (ECCV)}, 2018.

\bibitem[Bhat et~al.(2021)Bhat, Danelljan, Yu, Gool, and Timofte]{bhat2021deep}
Goutam Bhat, Martin Danelljan, Fisher Yu, Luc~Van Gool, and Radu Timofte.
\newblock Deep reparametrization of multi-frame super-resolution and denoising, 2021.

\bibitem[Bogoni(2000)]{bogoni2000extending}
Luca Bogoni.
\newblock Extending dynamic range of monochrome and color images through fusion.
\newblock In \emph{Proceedings 15th International Conference on Pattern Recognition. ICPR-2000}, pages 7--12. IEEE, 2000.

\bibitem[Brooks et~al.(2019)Brooks, Mildenhall, Xue, Chen, Sharlet, and Barron]{brooks2019unprocessing}
Tim Brooks, Ben Mildenhall, Tianfan Xue, Jiawen Chen, Dillon Sharlet, and Jonathan~T Barron.
\newblock Unprocessing images for learned raw denoising.
\newblock In \emph{Proceedings of the IEEE/CVF conference on computer vision and pattern recognition}, pages 11036--11045, 2019.

\bibitem[Chang et~al.(2021)Chang, Feng, Xu, and Li]{chang2021low}
Meng Chang, Huajun Feng, Zhihai Xu, and Qi Li.
\newblock Low-light image restoration with short-and long-exposure raw pairs.
\newblock \emph{IEEE Transactions on Multimedia}, 24:\penalty0 702--714, 2021.

\bibitem[Cho et~al.(2021)Cho, Ji, Hong, Jung, and Ko]{cho2021rethinking}
Sung-Jin Cho, Seo-Won Ji, Jun-Pyo Hong, Seung-Won Jung, and Sung-Jea Ko.
\newblock Rethinking coarse-to-fine approach in single image deblurring.
\newblock In \emph{Proceedings of the IEEE/CVF international conference on computer vision}, pages 4641--4650, 2021.

\bibitem[Ding et~al.(2022)Ding, Zhang, Han, and Ding]{ding2022scaling}
Xiaohan Ding, Xiangyu Zhang, Jungong Han, and Guiguang Ding.
\newblock Scaling up your kernels to 31x31: Revisiting large kernel design in cnns.
\newblock In \emph{Proceedings of the IEEE/CVF conference on computer vision and pattern recognition}, pages 11963--11975, 2022.

\bibitem[Eilertsen et~al.(2017)Eilertsen, Kronander, Denes, Mantiuk, and Unger]{eilertsen2017hdr}
Gabriel Eilertsen, Joel Kronander, Gyorgy Denes, Rafa{\l}~K Mantiuk, and Jonas Unger.
\newblock Hdr image reconstruction from a single exposure using deep cnns.
\newblock \emph{ACM transactions on graphics (TOG)}, 36\penalty0 (6):\penalty0 1--15, 2017.

\bibitem[Froehlich et~al.(2014)Froehlich, Grandinetti, Eberhardt, Walter, Schilling, and Brendel]{froehlich2014creating}
Jan Froehlich, Stefan Grandinetti, Bernd Eberhardt, Simon Walter, Andreas Schilling, and Harald Brendel.
\newblock Creating cinematic wide gamut hdr-video for the evaluation of tone mapping operators and hdr-displays.
\newblock In \emph{Digital photography X}, pages 279--288. SPIE, 2014.

\bibitem[Godard et~al.(2017)Godard, Matzen, and Uyttendaele]{godard2017deep}
Clément Godard, Kevin Matzen, and Matt Uyttendaele.
\newblock Deep burst denoising, 2017.

\bibitem[Grosch et~al.(2006)]{grosch2006fast}
Thorsten Grosch et~al.
\newblock Fast and robust high dynamic range image generation with camera and object movement.
\newblock \emph{Vision, Modeling and Visualization, RWTH Aachen}, 277284\penalty0 (3):\penalty0 2, 2006.

\bibitem[Guo et~al.(2019)Guo, Yan, Zhang, Zuo, and Zhang]{guo2019toward}
Shi Guo, Zifei Yan, Kai Zhang, Wangmeng Zuo, and Lei Zhang.
\newblock Toward convolutional blind denoising of real photographs.
\newblock In \emph{Proceedings of the IEEE/CVF conference on computer vision and pattern recognition}, pages 1712--1722, 2019.

\bibitem[Hu et~al.(2013)Hu, Gallo, Pulli, and Sun]{hu2013hdr}
Jun Hu, Orazio Gallo, Kari Pulli, and Xiaobai Sun.
\newblock Hdr deghosting: How to deal with saturation?
\newblock In \emph{Proceedings of the IEEE conference on computer vision and pattern recognition}, pages 1163--1170, 2013.

\bibitem[Hu et~al.(2024)Hu, Yan, Qi, and Zhang]{hu2024generating}
Tao Hu, Qingsen Yan, Yuankai Qi, and Yanning Zhang.
\newblock Generating content for hdr deghosting from frequency view.
\newblock \emph{arXiv preprint arXiv:2404.00849}, 2024.

\bibitem[Kalantari et~al.(2017)Kalantari, Ramamoorthi, et~al.]{kalantari2017deep}
Nima~Khademi Kalantari, Ravi Ramamoorthi, et~al.
\newblock Deep high dynamic range imaging of dynamic scenes.
\newblock \emph{ACM Trans. Graph.}, 36\penalty0 (4):\penalty0 144--1, 2017.

\bibitem[Kang et~al.(2003)Kang, Uyttendaele, Winder, and Szeliski]{kang2003high}
Sing~Bing Kang, Matthew Uyttendaele, Simon Winder, and Richard Szeliski.
\newblock High dynamic range video.
\newblock \emph{ACM Transactions on Graphics (TOG)}, 22\penalty0 (3):\penalty0 319--325, 2003.

\bibitem[Kim and Kim(2023)]{kim2023joint}
Jungwoo Kim and Min~H Kim.
\newblock Joint demosaicing and deghosting of time-varying exposures for single-shot hdr imaging.
\newblock In \emph{Proceedings of the IEEE/CVF International Conference on Computer Vision}, pages 12292--12301, 2023.

\bibitem[Lai et~al.(2022)Lai, Shih, Chu, Wu, Tsai, Krainin, Sun, and Liang]{lai2022face}
Wei-Sheng Lai, Yichang Shih, Lun-Cheng Chu, Xiaotong Wu, Sung-Fang Tsai, Michael Krainin, Deqing Sun, and Chia-Kai Liang.
\newblock Face deblurring using dual camera fusion on mobile phones.
\newblock \emph{ACM Transactions on Graphics (TOG)}, 41\penalty0 (4):\penalty0 1--16, 2022.

\bibitem[Lecouat et~al.(2022)Lecouat, Eboli, Ponce, and Mairal]{lecouat2022high}
Bruno Lecouat, Thomas Eboli, Jean Ponce, and Julien Mairal.
\newblock High dynamic range and super-resolution from raw image bursts.
\newblock \emph{arXiv preprint arXiv:2207.14671}, 2022.

\bibitem[Li et~al.(2022)Li, Wang, Wang, Liu, Zhang, Zeng, and Liu]{9826814}
Ru Li, Chuan Wang, Jue Wang, Guanghui Liu, Heng-Yu Zhang, Bing Zeng, and Shuaicheng Liu.
\newblock Uphdr-gan: Generative adversarial network for high dynamic range imaging with unpaired data.
\newblock \emph{IEEE Transactions on Circuits and Systems for Video Technology}, 32\penalty0 (11):\penalty0 7532--7546, 2022.

\bibitem[Li et~al.(2023)Li, Zhang, Timofte, Van~Gool, Tu, Du, Wang, Chen, Li, Wang, et~al.]{li2023ntire}
Yawei Li, Yulun Zhang, Radu Timofte, Luc Van~Gool, Zhijun Tu, Kunpeng Du, Hailing Wang, Hanting Chen, Wei Li, Xiaofei Wang, et~al.
\newblock Ntire 2023 challenge on image denoising: Methods and results.
\newblock In \emph{Proceedings of the IEEE/CVF Conference on Computer Vision and Pattern Recognition}, pages 1904--1920, 2023.

\bibitem[Liu et~al.(2022{\natexlab{a}})Liu, Mao, Wu, Feichtenhofer, Darrell, and Xie]{liu2022convnet}
Zhuang Liu, Hanzi Mao, Chao-Yuan Wu, Christoph Feichtenhofer, Trevor Darrell, and Saining Xie.
\newblock A convnet for the 2020s.
\newblock In \emph{Proceedings of the IEEE/CVF conference on computer vision and pattern recognition}, pages 11976--11986, 2022{\natexlab{a}}.

\bibitem[Liu et~al.(2022{\natexlab{b}})Liu, Wang, Zeng, and Liu]{liu2022ghost}
Zhen Liu, Yinglong Wang, Bing Zeng, and Shuaicheng Liu.
\newblock Ghost-free high dynamic range imaging with context-aware transformer.
\newblock In \emph{European Conference on Computer Vision}, pages 344--360. Springer, 2022{\natexlab{b}}.

\bibitem[Lu et~al.(2022)Lu, Li, Liu, Huang, Zhang, and Zeng]{lu2022transformer}
Zhisheng Lu, Juncheng Li, Hong Liu, Chaoyan Huang, Linlin Zhang, and Tieyong Zeng.
\newblock Transformer for single image super-resolution.
\newblock In \emph{Proceedings of the IEEE/CVF conference on computer vision and pattern recognition}, pages 457--466, 2022.

\bibitem[Lucas and Kanade(1981)]{lucas1981iterative}
Bruce~D Lucas and Takeo Kanade.
\newblock An iterative image registration technique with an application to stereo vision.
\newblock In \emph{IJCAI'81: 7th international joint conference on Artificial intelligence}, pages 674--679, 1981.

\bibitem[Ma et~al.(2017)Ma, Li, Yong, Wang, Meng, and Zhang]{ma2017robust}
Kede Ma, Hui Li, Hongwei Yong, Zhou Wang, Deyu Meng, and Lei Zhang.
\newblock Robust multi-exposure image fusion: a structural patch decomposition approach.
\newblock \emph{IEEE Transactions on Image Processing}, 26\penalty0 (5):\penalty0 2519--2532, 2017.

\bibitem[Mao et~al.(2023)Mao, Liu, Liu, Li, Shen, and Wang]{mao2023intriguing}
Xintian Mao, Yiming Liu, Fengze Liu, Qingli Li, Wei Shen, and Yan Wang.
\newblock Intriguing findings of frequency selection for image deblurring.
\newblock In \emph{Proceedings of the AAAI Conference on Artificial Intelligence}, pages 1905--1913, 2023.

\bibitem[Mehta et~al.(2023)Mehta, Dudhane, Murala, Zamir, Khan, and Khan]{mehta2023gated}
Nancy Mehta, Akshay Dudhane, Subrahmanyam Murala, Syed~Waqas Zamir, Salman Khan, and Fahad~Shahbaz Khan.
\newblock Gated multi-resolution transfer network for burst restoration and enhancement.
\newblock In \emph{2023 IEEE/CVF Conference on Computer Vision and Pattern Recognition (CVPR)}, pages 22201--22210. IEEE, 2023.

\bibitem[Mildenhall et~al.(2018)Mildenhall, Barron, Chen, Sharlet, Ng, and Carroll]{mildenhall2018burst}
Ben Mildenhall, Jonathan~T. Barron, Jiawen Chen, Dillon Sharlet, Ren Ng, and Robert Carroll.
\newblock Burst denoising with kernel prediction networks, 2018.

\bibitem[Nah et~al.(2017)Nah, Hyun~Kim, and Mu~Lee]{nah2017deep}
Seungjun Nah, Tae Hyun~Kim, and Kyoung Mu~Lee.
\newblock Deep multi-scale convolutional neural network for dynamic scene deblurring.
\newblock In \emph{Proceedings of the IEEE conference on computer vision and pattern recognition}, pages 3883--3891, 2017.

\bibitem[Niu et~al.(2021)Niu, Wu, Liu, Guo, and Lau]{niu2021hdr}
Yuzhen Niu, Jianbin Wu, Wenxi Liu, Wenzhong Guo, and Rynson~WH Lau.
\newblock Hdr-gan: Hdr image reconstruction from multi-exposed ldr images with large motions.
\newblock \emph{IEEE Transactions on Image Processing}, 30:\penalty0 3885--3896, 2021.

\bibitem[Pece and Kautz(2010)]{pece2010bitmap}
Fabrizio Pece and Jan Kautz.
\newblock Bitmap movement detection: Hdr for dynamic scenes.
\newblock In \emph{2010 Conference on Visual Media Production}, pages 1--8. IEEE, 2010.

\bibitem[Prabhakar et~al.(2019)Prabhakar, Arora, Swaminathan, Singh, and Babu]{prabhakar2019fast}
K~Ram Prabhakar, Rajat Arora, Adhitya Swaminathan, Kunal~Pratap Singh, and R~Venkatesh Babu.
\newblock A fast, scalable, and reliable deghosting method for extreme exposure fusion.
\newblock In \emph{2019 IEEE International Conference on Computational Photography (ICCP)}, pages 1--8. IEEE, 2019.

\bibitem[Prabhakar et~al.(2020)Prabhakar, Agrawal, Singh, Ashwath, and Babu]{prabhakar2020towards}
K~Ram Prabhakar, Susmit Agrawal, Durgesh~Kumar Singh, Balraj Ashwath, and R~Venkatesh Babu.
\newblock Towards practical and efficient high-resolution hdr deghosting with cnn.
\newblock In \emph{Computer Vision--ECCV 2020: 16th European Conference, Glasgow, UK, August 23--28, 2020, Proceedings, Part XXI 16}, pages 497--513. Springer, 2020.

\bibitem[Ranjan and Black(2017)]{ranjan2017optical}
Anurag Ranjan and Michael~J Black.
\newblock Optical flow estimation using a spatial pyramid network.
\newblock In \emph{Proceedings of the IEEE conference on computer vision and pattern recognition}, pages 4161--4170, 2017.

\bibitem[Sen et~al.(2012)Sen, Kalantari, Yaesoubi, Darabi, Goldman, and Shechtman]{sen2012robust}
Pradeep Sen, Nima~Khademi Kalantari, Maziar Yaesoubi, Soheil Darabi, Dan~B Goldman, and Eli Shechtman.
\newblock Robust patch-based hdr reconstruction of dynamic scenes.
\newblock \emph{ACM Trans. Graph.}, 31\penalty0 (6):\penalty0 203--1, 2012.

\bibitem[Shekarforoush et~al.(2023)Shekarforoush, Walia, Brubaker, Derpanis, and Levinshtein]{shekarforoush2023dual}
Shayan Shekarforoush, Amanpreet Walia, Marcus~A Brubaker, Konstantinos~G Derpanis, and Alex Levinshtein.
\newblock Dual-camera joint deblurring-denoising.
\newblock \emph{arXiv preprint arXiv:2309.08826}, 2023.

\bibitem[Song et~al.(2022)Song, Park, Kong, Kwak, and Kang]{song2022selective}
Jou~Won Song, Ye-In Park, Kyeongbo Kong, Jaeho Kwak, and Suk-Ju Kang.
\newblock Selective transhdr: Transformer-based selective hdr imaging using ghost region mask.
\newblock In \emph{European Conference on Computer Vision}, pages 288--304. Springer, 2022.

\bibitem[Tao et~al.(2018)Tao, Gao, Shen, Wang, and Jia]{tao2018scale}
Xin Tao, Hongyun Gao, Xiaoyong Shen, Jue Wang, and Jiaya Jia.
\newblock Scale-recurrent network for deep image deblurring.
\newblock In \emph{Proceedings of the IEEE conference on computer vision and pattern recognition}, pages 8174--8182, 2018.

\bibitem[Tel et~al.(2023)Tel, Wu, Zhang, Heyrman, Demonceaux, Timofte, and Ginhac]{tel2023alignment}
Steven Tel, Zongwei Wu, Yulun Zhang, Barth{\'e}l{\'e}my Heyrman, C{\'e}dric Demonceaux, Radu Timofte, and Dominique Ginhac.
\newblock Alignment-free hdr deghosting with semantics consistent transformer.
\newblock \emph{arXiv preprint arXiv:2305.18135}, 2023.

\bibitem[Tian et~al.(2024)Tian, Zhang, Li, Yao, and Pan]{tian2024multi}
Huiyuan Tian, Li Zhang, Shijian Li, Min Yao, and Gang Pan.
\newblock Multi-depth branch network for efficient image super-resolution.
\newblock \emph{Image and Vision Computing}, page 104949, 2024.

\bibitem[Tomaszewska and Mantiuk(2007)]{tomaszewska2007image}
Anna Tomaszewska and Radoslaw Mantiuk.
\newblock Image registration for multi-exposure high dynamic range image acquisition.
\newblock 2007.

\bibitem[Wei et~al.(2023)Wei, Sun, Guo, Liu, Li, Chen, Ji, and Lin]{wei2023towards}
Pengxu Wei, Yujing Sun, Xingbei Guo, Chang Liu, Guanbin Li, Jie Chen, Xiangyang Ji, and Liang Lin.
\newblock Towards real-world burst image super-resolution: Benchmark and method.
\newblock In \emph{Proceedings of the IEEE/CVF International Conference on Computer Vision}, pages 13233--13242, 2023.

\bibitem[Wu et~al.(2020)Wu, Liu, Cao, Ren, and Zuo]{wu2020unpaired}
Xiaohe Wu, Ming Liu, Yue Cao, Dongwei Ren, and Wangmeng Zuo.
\newblock Unpaired learning of deep image denoising.
\newblock In \emph{European conference on computer vision}, pages 352--368. Springer, 2020.

\bibitem[Yan et~al.(2019)Yan, Gong, Shi, Hengel, Shen, Reid, and Zhang]{yan2019attention}
Qingsen Yan, Dong Gong, Qinfeng Shi, Anton van~den Hengel, Chunhua Shen, Ian Reid, and Yanning Zhang.
\newblock Attention-guided network for ghost-free high dynamic range imaging.
\newblock In \emph{Proceedings of the IEEE/CVF Conference on Computer Vision and Pattern Recognition}, pages 1751--1760, 2019.

\bibitem[Yan et~al.(2020)Yan, Zhang, Liu, Zhu, Sun, Shi, and Zhang]{yan2020deep}
Qingsen Yan, Lei Zhang, Yu Liu, Yu Zhu, Jinqiu Sun, Qinfeng Shi, and Yanning Zhang.
\newblock Deep hdr imaging via a non-local network.
\newblock \emph{IEEE Transactions on Image Processing}, 29:\penalty0 4308--4322, 2020.

\bibitem[Yan et~al.(2022)Yan, Gong, Shi, van~den Hengel, Shen, Reid, and Zhang]{yan2022dual}
Qingsen Yan, Dong Gong, Javen~Qinfeng Shi, Anton van~den Hengel, Chunhua Shen, Ian Reid, and Yanning Zhang.
\newblock Dual-attention-guided network for ghost-free high dynamic range imaging.
\newblock \emph{International Journal of Computer Vision}, pages 1--19, 2022.

\bibitem[Yan et~al.(2023{\natexlab{a}})Yan, Chen, Zhang, Zhu, Sun, and Zhang]{yan2023unified}
Qingsen Yan, Weiye Chen, Song Zhang, Yu Zhu, Jinqiu Sun, and Yanning Zhang.
\newblock A unified hdr imaging method with pixel and patch level.
\newblock In \emph{Proceedings of the IEEE/CVF Conference on Computer Vision and Pattern Recognition}, pages 22211--22220, 2023{\natexlab{a}}.

\bibitem[Yan et~al.(2023{\natexlab{b}})Yan, Hu, Sun, Tang, Zhu, Dong, Van~Gool, and Zhang]{yan2023towards}
Qingsen Yan, Tao Hu, Yuan Sun, Hao Tang, Yu Zhu, Wei Dong, Luc Van~Gool, and Yanning Zhang.
\newblock Towards high-quality hdr deghosting with conditional diffusion models.
\newblock \emph{IEEE Transactions on Circuits and Systems for Video Technology}, 2023{\natexlab{b}}.

\bibitem[Yan et~al.(2023{\natexlab{c}})Yan, Zhang, Chen, Tang, Zhu, Sun, Van~Gool, and Zhang]{yan2023smae}
Qingsen Yan, Song Zhang, Weiye Chen, Hao Tang, Yu Zhu, Jinqiu Sun, Luc Van~Gool, and Yanning Zhang.
\newblock Smae: Few-shot learning for hdr deghosting with saturation-aware masked autoencoders.
\newblock In \emph{Proceedings of the IEEE/CVF Conference on Computer Vision and Pattern Recognition}, pages 5775--5784, 2023{\natexlab{c}}.

\bibitem[Yuan et~al.(2007)Yuan, Sun, Quan, and Shum]{yuan2007image}
Lu Yuan, Jian Sun, Long Quan, and Heung-Yeung Shum.
\newblock Image deblurring with blurred/noisy image pairs.
\newblock In \emph{ACM SIGGRAPH 2007 papers}, pages 1--es. 2007.

\bibitem[Zamir et~al.(2020)Zamir, Arora, Khan, Hayat, Khan, Yang, and Shao]{zamir2020cycleisp}
Syed~Waqas Zamir, Aditya Arora, Salman Khan, Munawar Hayat, Fahad~Shahbaz Khan, Ming-Hsuan Yang, and Ling Shao.
\newblock Cycleisp: Real image restoration via improved data synthesis.
\newblock In \emph{Proceedings of the IEEE/CVF conference on computer vision and pattern recognition}, pages 2696--2705, 2020.

\bibitem[Zhang et~al.(2017)Zhang, Zuo, Chen, Meng, and Zhang]{zhang2017beyond}
Kai Zhang, Wangmeng Zuo, Yunjin Chen, Deyu Meng, and Lei Zhang.
\newblock Beyond a gaussian denoiser: Residual learning of deep cnn for image denoising.
\newblock \emph{IEEE transactions on image processing}, 26\penalty0 (7):\penalty0 3142--3155, 2017.

\bibitem[Zhang and Cham(2011)]{zhang2011gradient}
Wei Zhang and Wai-Kuen Cham.
\newblock Gradient-directed multiexposure composition.
\newblock \emph{IEEE Transactions on Image Processing}, 21\penalty0 (4):\penalty0 2318--2323, 2011.

\bibitem[Zhang et~al.(2022)Zhang, Xu, Liu, Yan, and Zuo]{zhang2022self}
Zhilu Zhang, RongJian Xu, Ming Liu, Zifei Yan, and Wangmeng Zuo.
\newblock Self-supervised image restoration with blurry and noisy pairs.
\newblock \emph{Advances in Neural Information Processing Systems}, 35:\penalty0 29179--29191, 2022.

\bibitem[Zhang et~al.(2024{\natexlab{a}})Zhang, Zhang, Wu, Yan, and Zuo]{zhang2024bracketing}
Zhilu Zhang, Shuohao Zhang, Renlong Wu, Zifei Yan, and Wangmeng Zuo.
\newblock Bracketing is all you need: Unifying image restoration and enhancement tasks with multi-exposure images.
\newblock \emph{arXiv preprint arXiv:2401.00766}, 2024{\natexlab{a}}.

\bibitem[Zhang et~al.(2024{\natexlab{b}})Zhang, Zhang, Wu, Zuo, Timofte, et~al.]{zhang2024ntirebrack}
Zhilu Zhang, Shuohao Zhang, Renlong Wu, Wangmeng Zuo, Radu Timofte, et~al.
\newblock Ntire 2024 challenge on bracketing image restoration and enhancement: Datasets, methods and results.
\newblock In \emph{Proceedings of the IEEE/CVF Conference on Computer Vision and Pattern Recognition (CVPR) Workshops}, 2024{\natexlab{b}}.

\bibitem[Zhu(2022)]{thop}
Ligeng Zhu.
\newblock Thop: Pytorch-opcounter, 2022.
\newblock \url{https://pypi.org/project/thop}.

\end{thebibliography}
}
\end{document}